\documentclass[10pt,twocolumn,letterpaper]{article}

\usepackage{cvpr}              

\usepackage{graphicx}
\usepackage{times}
\usepackage{helvet}
\usepackage{courier}
\usepackage{amsmath}
\usepackage{algorithm}
\usepackage{algorithmic}
\usepackage{csquotes} 
\usepackage{color}
\usepackage{paralist}
\usepackage{amssymb}
\usepackage{indentfirst}
\usepackage{float}
\usepackage{multirow}
\usepackage{cite}
\usepackage{mathrsfs}

\usepackage{mathrsfs}
\usepackage{textcomp,booktabs}
\usepackage{amssymb}
\usepackage{pifont}
\usepackage[misc]{ifsym}
\newcommand{\cmark}{\ding{51}}%
\newcommand{\xmark}{\ding{55}}%

\usepackage{mathrsfs}
\usepackage{color}
\usepackage{xcolor}
\definecolor{citecolor}{HTML}{0071bc} 
\definecolor{SeaGreen4}{RGB}{0,205,102} 
\definecolor{SlateBlue}{RGB}{106,90,205} 
\definecolor{DarkRed}{RGB}{178,34,34} 
\usepackage[colorlinks, linkcolor=red,  anchorcolor=blue, citecolor=citecolor]{hyperref}

\usepackage{colortbl}
\definecolor{mygray}{gray}{.9}
\definecolor{mypink}{rgb}{.99,.91,.95}
\definecolor{mycyan}{cmyk}{.3,0,0,0}

\usepackage{color}
\usepackage{xcolor}
\definecolor{citecolor}{HTML}{0071bc} 
\definecolor{SeaGreen4}{RGB}{0,205,102} 
\definecolor{SlateBlue}{RGB}{106,90,205} 
\definecolor{DarkRed}{RGB}{178,34,34}

\usepackage[capitalize]{cleveref}
\crefname{section}{Sec.}{Secs.}
\Crefname{section}{Section}{Sections}
\Crefname{table}{Table}{Tables}
\crefname{table}{Tab.}{Tabs.}

\def\cvprPaperID{*****} 
\def\confName{CVPR}
\def\confYear{2023}

\begin{document}

\title{ Event Stream-based Visual Object Tracking: A High-Resolution Benchmark Dataset and A Novel Baseline }


\author{Xiao Wang$^{1}$, Shiao Wang$^{1}$, Chuanming Tang$^{2,3}$, Lin Zhu$^{4}$, Bo Jiang  $^{1}$\thanks{\Letter~~Corresponding Author: Bo Jiang}, Yonghong Tian$^{5,6}$, Jin Tang$^{1}$ \\ 
${^1}$\emph{School of Computer Science and Technology, Anhui University, Hefei, China} \\
${^2}$\emph{University of Chinese Academy of Sciences, Beijing, China} \\ 
${^3}$\emph{Institute of Optics and Electronics, CAS, Chengdu, China} \\ 
${^4}$\emph{Beijing Institute of Technology, Beijing, China} \\
${^5}$\emph{Peng Cheng Laboratory, Shenzhen, China} ~~~
${^6}$\emph{Peking University, Beijing, China} }

\maketitle

\begin{abstract}
Tracking using bio-inspired event cameras has drawn more and more attention in recent years. Existing works either utilize aligned RGB and event data for accurate tracking or directly learn an event-based tracker. The first category needs more cost for inference and the second one may be easily influenced by noisy events or sparse spatial resolution. In this paper, we propose a novel hierarchical knowledge distillation framework that can fully utilize multi-modal / multi-view information during training to facilitate knowledge transfer, enabling us to achieve high-speed and low-latency visual tracking during testing by using only event signals. Specifically, a teacher Transformer-based multi-modal tracking framework is first trained by feeding the RGB frame and event stream simultaneously. Then, we design a new hierarchical knowledge distillation strategy which includes pairwise similarity, feature representation, and response maps-based knowledge distillation to guide the learning of the student Transformer network. Moreover, since existing event-based tracking datasets are all low-resolution ($346 \times 260$), we propose the first large-scale high-resolution ($1280 \times 720$) dataset named EventVOT. It contains 1141 videos and covers a wide range of categories such as pedestrians, vehicles, UAVs, ping pongs, etc. Extensive experiments on both low-resolution (FE240hz, VisEvent, COESOT), and our newly proposed high-resolution EventVOT dataset fully validated the effectiveness of our proposed method. The dataset, evaluation toolkit, and source code are available on \url{https://github.com/Event-AHU/EventVOT_Benchmark}  
\end{abstract}

\section{Introduction} 

Visual Object Tracking (VOT) targets predicting the locations of target object initialized in the first frame. Existing trackers are usually developed based on RGB cameras and deployed for autonomous driving, drone photography, intelligent video surveillance and other fields. Due to the influence of challenging factors like fast motion, illumination, background distractor, and out-of-view, the tracking performance in complex scenarios is still unsatisfactory. The video frames with these challenges are unevenly distributed in the tracking video, making it difficult to improve the overall tracking results by investing more labelled data.

\begin{figure*}[!htp]
\center
\includegraphics[width=7in]{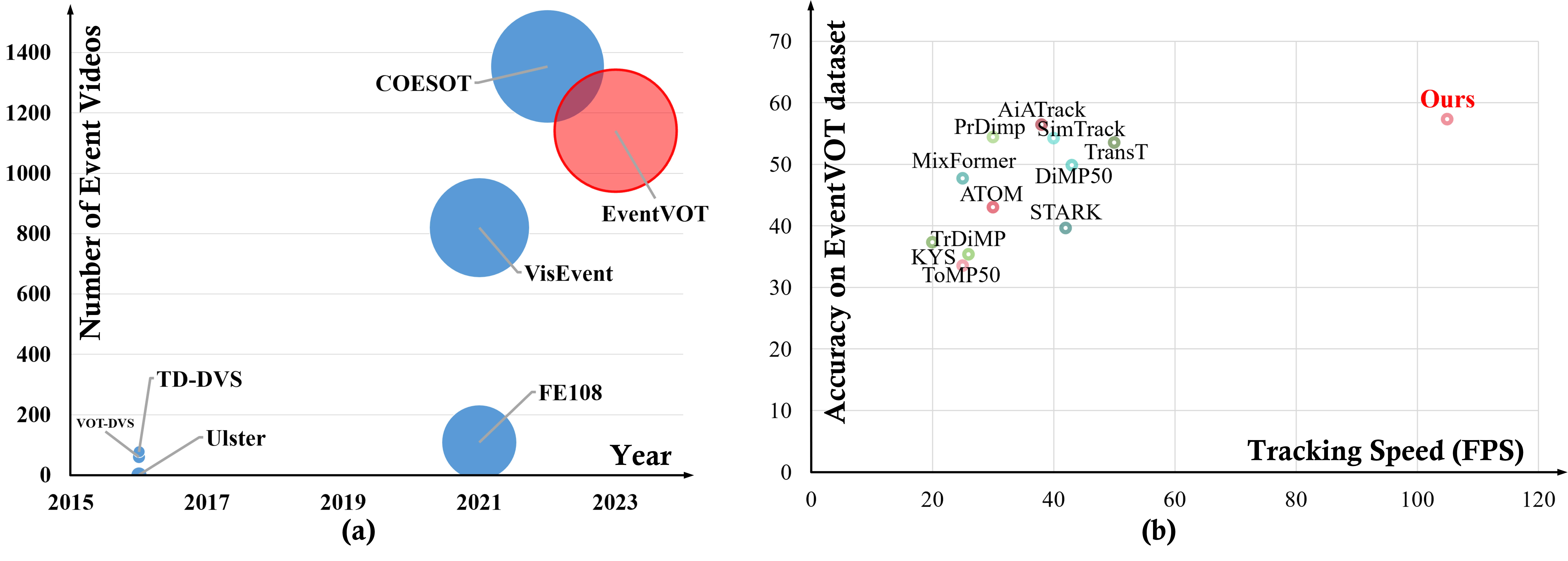}
\caption{ (a). Comparison between our newly proposed EventVOT and other event-based tracking datasets; 
(b). Comparison between our tracker and existing SOTA trackers on the tracking speed and accuracy on the EventVOT dataset.} 
\label{firstIMG}
\end{figure*}

To address these challenges, some researchers have started to improve the effectiveness of input data by introducing new sensors. As a new type of bio-inspired sensor, event cameras are different from traditional video frame sensors in that they can output event pulses asynchronously and capture motion information through the detection of events (e.g., changes in light intensity). Event camera performs better than traditional RGB cameras in capturing fast-moving objects due to dense temporal resolution. It also works well on high dynamic range, low energy consumption, and low latency~\cite{gallego2020event}. Event cameras can be used for a wide range of applications, including surveillance, robotics, medical imaging, and sports analysis.

Although few, there have been some studies that exploit event cameras for visual object tracking. For example, Zhang et al. propose  AFNet~\cite{zhang2023AFNet} and CDFI~\cite{zhang2021fe108} to combine the frame and event data via multi-modality alignment and fusion modules. STNet~\cite{zhang2022STN} is proposed to connect the Transformer and spiking neural networks for event-based tracking. Zhu et al.~\cite{zhu2022grapheventTrack} attempt to mine the key events and employ a graph-based network to embed the irregular spatio-temporal information of key events into a high-dimensional feature space for tracking. These works attempt to obtain stronger tracking algorithms through multi-modal fusion or pure event training and tracking methods. Although good performance can be achieved, however, these algorithms are still easily influenced by the following issues:
\textbf{Firstly}, the spatial signal of event cameras is very sparse in slow-moving scenes, and the contours of target objects are not clear enough, which may lead to tracking failures. Tracking using RGB-Event data can better compensate for this deficiency, but additional modalities will increase the cost of model inference. 
\textbf{Secondly}, existing event-based tracking datasets are collected using the DVS346 camera, which has an output resolution of $346 \times 260$. It has not been explored or validated whether the event representation and feature extraction methods designed for low-resolution event stream are still effective on high-resolution event data. 
Therefore, it is natural to raise the following open question: \emph{Can we transfer knowledge from multi-modal or multi-view data during the training phase and achieve robust tracking by only using the event data during the testing phase?}

\begin{figure*}[!htp]
\center
\includegraphics[width=6in]{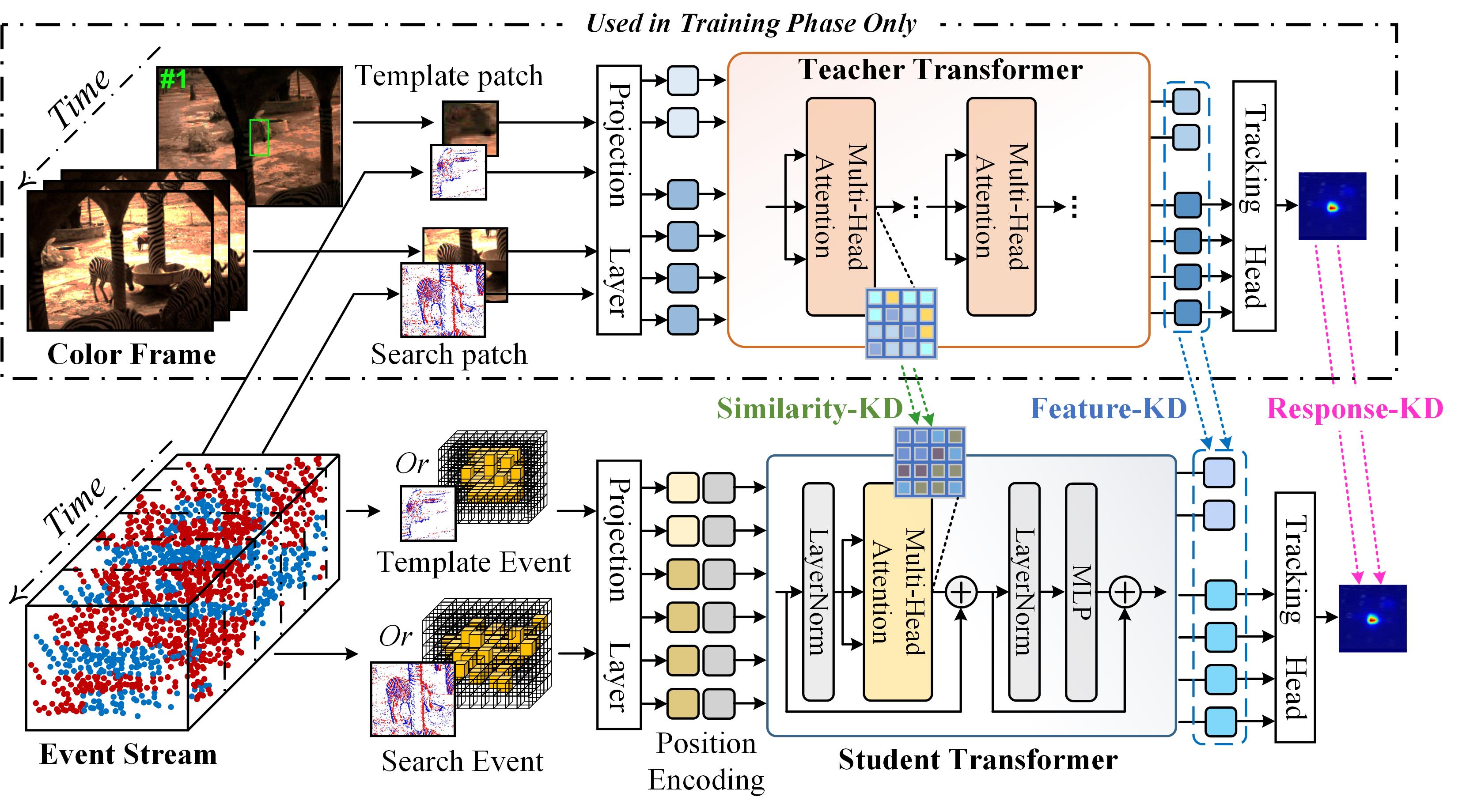}
\caption{\textbf{An overview of our proposed Hierarchical Knowledge Distill Framework for Event Stream based Tracking, termed HDETrack.} It contains the teacher and student Transformer network which takes multi-modal/multi-view and event data only as the input. The two networks share the same architecture, i.e., tracking using a unified Transformer backbone network similar to OSTrack~\cite{ye2022Ostrack} and CEUTrack~\cite{tang2022coesot}. Specifically, we extract the template and search patches of RGB and event input and obtain feature embeddings using a projection layer. Then, a couple of Transformer layers are stacked as the teacher network. The output will be fed into the tracking head for target object localization. Meanwhile, the student takes event data only for low latency tracking which is trained using tracking loss functions and also knowledge distillation from teacher Transformer networks. Our tracker achieves a better tradeoff between accuracy and model complexity, as shown in Fig.~\ref{firstIMG}. } 
\label{framework}
\end{figure*}

In this work, we propose a novel event-based visual tracking framework by designing a new cross-modality hierarchical knowledge distillation scheme. As shown in Fig.~\ref{framework}, we first train a \textbf{teacher} Transformer network by feeding the RGB frame and event stream. It crops the template patch and search region of dual-modality from the initialized and subsequent frames respectively and adopts a projection layer to transform them into token representations. Then, a couple of Transformer blocks are used to fuse the tokens as a unified backbone. Finally, the tracking head is adopted to predict the response maps for target localization. Once we obtain the teacher Transformer network, the hierarchical knowledge distillation strategy is conducted to guide the learning of the \textbf{student} Transformer network which only the event data are fed. To be specific, the similarity matrix, feature representation, and response maps based knowledge distillation are simultaneously considered for cross-modality knowledge transfer. Note that, since only the event data are fed in the student network, it can achieve not only accurate but also low-latency and high-speed object tracking in the testing stage.

Moreover, in addition to evaluating our tracker on existing event-based tracking datasets, we also propose a new high-resolution event-based tracking dataset, termed EventVOT, to fully validate the effectiveness of our method and other related works. Different from existing datasets with limited resolution (e.g., FE240hz, VisEvent, COESOT are $346 \times 260$) as shown in Fig.~\ref{firstIMG} (a), our videos are collected using the Prophesee camera EVK4–HD which outputs event stream in $1280 \times 720$. It contains 1141 videos and covers a wide range of target objects, including pedestrians, vehicles, UAVs, ping pongs, etc. To build a comprehensive benchmark dataset, we provide the tracking results of multiple baseline trackers for future work to compare. We hope our newly proposed EventVOT dataset can open up new possibilities for event tracking.

To sum up, our contributions can be concluded as the following three aspects: 
\begin{itemize}
    \item  We propose a novel hierarchical cross-modality knowledge distillation strategy for event-based tracking. It is the first work to exploit the knowledge transfer from multi-modal (RGB-Event) / multi-view (Event Image-Voxel) to an unimodal event-based tracker, termed HDETrack.
    \item We propose the first high-resolution benchmark dataset for event-based tracking, termed EventVOT. We also provide experimental evaluations of recent strong trackers to build a comprehensive event-based tracking benchmark. 
    \item Extensive experiments on four large-scale benchmark datasets, i.e., FE240hz, VisEvent, COESOT, and EventVOT, fully validate the effectiveness of our proposed tracker. 
\end{itemize}

\section{Related Work} 

In this section, we review the most related research topics to our paper, including RGB camera based tracking, event camera based tracking, and knowledge distillation. More related works can be found in the following surveys~\cite{gou2021KDSurvey, gallego2020event, marvasti2021trackSurvey} and paper list \footnote{\url{github.com/wangxiao5791509/Single_Object_Tracking_Paper_List}}.

\noindent 
\textbf{RGB Camera based Tracking. } 
The mainstream visual trackers are developed based on RGB videos and boosted by deep learning techniques in recent years. The convolutional neural networks are first adopted for feature extraction and learning. Specifically, the MDNet series~\cite{hyeonseob2016MDNet} extract the deep features using three convolutional layers and learn domain-specific layers for tracking. The SiamFC~\cite{luca2016SiamFC} and SINT~\cite{ran2016SINT} first utilize the Siamese fully convolutional neural networks and Siamese instance matching for tracking, respectively. Later, the Siamese network based trackers become the mainstream gradually and many representative trackers are proposed, like SiamRPN++~\cite{li2019SiamRPN++}, SiamMask~\cite{wang2019Siammask}, SiamBAN~\cite{chen2020SIamBAN}, Ocean~\cite{zhang2020Ocean}, LTM~\cite{zhang2021LTM}, ATOM~\cite{martin2019Atom}, DiMP~\cite{goutam2019Dimp}, PrDiMP~\cite{martin2020PrDimp}, etc.

Inspired by the success of self-attention and Transformer networks in natural language processing, some researchers also exploit Transformers for visual object tracking~\cite{wang2021TrDiMP, chen2021transt, mayer2022Tomp, gao2022AIa, chen2022SimTrack, ye2022Ostrack, zhang2022STN, tang2022coesot}. For example, Wang et al.~\cite{wang2021TrDiMP} proposed TrDiMP, which integrates Transformer with tracking tasks, exploits temporal context for robust visual tracking. Chen et al.~\cite{chen2021transt} proposed TransT, a new attention-based feature fusion network and a Siamese structured tracking approach that integrates a fusion network have been designed using Transformer. Other works like Tomp~\cite{mayer2022Tomp} proposed a Transformer-based model prediction module, enabling it to learn more powerful target prediction capabilities, due to the powerful inductive bias of Transformer in capturing global relationships. Gao et al.~\cite{gao2022AIa} proposed AiATrack that introduce a universal feature extraction and information propagation module based on Transformer. A simplified tracking architecture called SimTrack~\cite{chen2022SimTrack} has been proposed by Chen et al. which utilize a Transformer as the backbone for joint feature extraction and interaction. Ye et al.~\cite{ye2022Ostrack} propose OSTrack, they design a one-stream tracking framework to replace the complex dual-stream framework. Zhang et al.~\cite{zhang2022STN} combine spiking neural networks with Transformer for event-based tracking. CEUTrack~\cite{tang2022coesot} is proposed by Tang et al., who explore a Transformer-based dual-modal framework for RGB-Event tracking. 
Different from these works, we exploit event cameras to achieve reliable tracking even under challenging scenarios, like low illumination and fast motion.

\noindent 
\textbf{Event Camera based Tracking. } 
Tracking using event cameras is a newly arising research topic and draws more and more attention in recent years. Specifically, early event-based trackers ESVM (event-guided support vector machine)~\cite{huang2018event} is proposed by Huang et al. for high-speed moving object tracking. AFNet~\cite{zhang2023AFNet} proposed by Zhang et al. incorporates event-guided cross-modality alignment (ECA) and cross-correlation fusion (CF) module, which effectively align and fuse the RGB and event streams. Chen et al.~\cite{chen2019asynchronous} propose an Adaptive Time-Surface with Linear Time Decay (ATSLTD) event-to-frame conversion algorithm for asynchronous retinal events based tracking. Representative works like EKLT~\cite{gehrig2020eklt} fuse the frame and event stream to track visual features with high temporal resolution. Zhang et al.~\cite{zhang2021fe108} adopt self- and cross-domain attention schemes to enhance the RGB and event features for robust tracking. STNet~\cite{zhang2022STN} is proposed to capture the global spatial information using Transformer and temporal cues using a spiking neural network (SNN). Wang et al.~\cite{wang2021visevent} fuse the RGB and event data using cross-modality Transformer module. Zhu et al.~\cite{zhu2022grapheventTrack} sample the key-events using a density-insensitive downsampling strategy and embed them into high-dimensional feature space for tracking via a graph-based network. Tang et al.~\cite{tang2022coesot} conduct RGB-Event tracking through a unified backbone network to simultaneously realize multimodal feature extraction, correlation, and fusion. Zhu et al.~\cite{zhu2023promptTrack} introduce prompt tuning to drive the pre-trained RGB backbone for multimodal tracking. AFNet~\cite{zhang2023AFNet} is proposed to combine meaningful information from both modalities at different measurement rates using multi-modality alignment and fusion modules. Zhu et al.~\cite{zhu2023crossRGBETrack} randomly masks tokens of a specific modality and proposes an orthogonal high-rank loss function to enforce the interaction between different modalities. 
Different from existing works, in this paper, we propose to conduct knowledge distill from multi-modal or multi-view in the training phase and only use event data for efficient and low-latency tracking.

\noindent 
\textbf{Knowledge Distillation. } 
Learning a student network using knowledge distillation for efficient and accurate inference is a widely studied problem. 
Deng et al.~\cite{deng2021distEvent} provide explicit feature-level supervision for the learning of event stream by using knowledge distilled from the image domain. 
For the tracking task, Shen et al.~\cite{shen2021distilledSiamTrack} propose to distill large Siamese trackers using a teacher-students knowledge distillation model for small, fast, and accurate trackers. Chen et al.~\cite{chen2022TSKDcorrTrack} attempt to learn a lightweight student correlation filter-based tracker by distilling a pre-trained deep convolutional neural network. Zhuang et al.~\cite{zhuang2021ensembleSiamTrack} introduce Ensemble learning (EL) into the Siamese tracking framework and treat two Siamese networks as students and enabling them to learn collaboratively. Zhang et al.~\cite{zhang2023RGBTDist} propose a specific-common feature distillation strategy to transform the modality-common and modality-specific information from a deeper two-stream network to a shallower single-stream network. Sun et al.~\cite{sun2021USLCMTrack} conduct cross-modal distillation for TIR tracking from RGB modality on unlabeled paired RGB-TIR data. Wang et al.~\cite{wang2020CFTrack} distill the CNN model pre-trained from the image classification dataset into a lightweight student network for fast correlation filter trackers. Zhao et al.~\cite{zhao2022distSiamTrack} propose a distillation-ensemble-selection framework to address the conflict between the tracking efficiency and model complexity. Ge et al.~\cite{ge2019distChannelTrack} propose channel distillation for correlation filter trackers which can accurately mine better channels and alleviate the influence of noisy channels. 
Different from these works, our proposed hierarchical knowledge distillation enables message propagation from multi-modality or multi-view to event-tracking networks.

\section{Methodology}  

In this section, we will introduce our proposed hierarchical knowledge distillation framework that helps the learning of event-based student tracker from multi-modal or multi-view teacher network. We will first give an overview of our proposed distillation framework, then, we introduce the RGB and event representations used in this work. After that, we explain the network architectures and hierarchical knowledge distillation strategies of the teacher-student framework. Finally, we summarize the loss functions used for the training of our network.

\subsection{Overview}  
To achieve efficient and low-latency visual tracking, in this paper, we exploit tracking using an event camera only. To ensure its tracking performance, we resort to the knowledge distillation (KD) from multi-modal or multi-view data. Therefore, we first train a large-scale teacher Transformer using the RGB frames and event stream, as shown in Fig.~\ref{framework}. To be specific, the template patch and search patch of dual modalities are extracted and transformed into tokens using the projection layer. These tokens are directly concatenated and fed into a unified Transformer backbone network for simultaneous feature extraction, interactive learning, and fusion. 
For the event student tracking network, we take the event images or voxels as the input and optimize the parameters based on tracking loss functions and knowledge distillation functions. More in detail, the similarity matrix based KD, feature based KD, and response map based KD are all considered for a higher tracking performance. We will introduce more details about the network architecture and hierarchical knowledge distillation strategies in the following subsections.

\subsection{Input Representation} 
In this work, we denote the RGB frames as $\mathcal{I} = \{I_1, I_2, ..., I_N\}$, where $I_i$ denotes each video frame, $i \in [1, N]$, $N$ is the number of video frames; and treat event stream as $\mathcal{E} = \{e_1, e_2, ..., e_M\}$, where $e_j$ denotes each event point asynchronously launched, $j \in [1, M]$, $M$ is the number of event points in current sample.

For the video frames $\mathcal{I}$, we utilize standard processing methods for Siamese tracking and extract the template patch $T_I$ and search patch $S_I$ as the input. 
For the event stream $\mathcal{E}$, we stack/split them into event images/voxels which can fuse more conveniently with existing RGB modality. More in detail, the event images are obtained by aligning with the exposure time of RGB modality. Event voxels are obtained by splitting the event stream along with the spatial (width $W$ and height $H$) and temporal dimensions ($T_i$). The scale of each voxel grid is denoted as ($a, b, c$), thus, we can get $\frac{W}{a} \times \frac{H}{b} \times \frac{T_i}{c}$ voxel grids. Similarly, we can obtain the template and search regions of event data, i.e., $T_E$ and $S_E$.

\subsection{Network Architecture} 

In this paper, we propose a novel hierarchical knowledge distillation framework for event-based tracking. As shown in Fig.~\ref{framework}, it mainly contains the Multimodal/Multiview Teacher Transformer and Unimodal Student Transformer network.

\noindent
\textbf{Multimodal/Multiview Teacher Tracker.~} 
We feed the RGB frame and event stream or different event data (e.g., event image and voxel) into the teacher Transformer network. The template and search patches of both modalities/views are concatenated and fed into a projection layer for feature embedding. Following the unified backbone based trackers~\cite{ye2022Ostrack, tang2022coesot}, we propose a teacher network consisting of Transformer layers for multimodal feature learning and fusion. Then, the tokens corresponding to the search region are selected for target object localization using the tracking head.

\noindent
\textbf{Unimodal Student Tracker.~} 
To achieve efficient and low-latency visual tracking, we don't conduct tracking using multimodal data. A lightweight student Transformer based tracker is proposed, as shown in Fig.~\ref{framework}. Note that, only event data is fed into the student Transformer for tracking. Due to the influence of challenging factors of event-based tracking, such as sparse event points, and clutter background, we introduce a hierarchical knowledge distillation strategy to enhance its tracking performance.

\subsection{Hierarchical Knowledge Distillation} 

The tracking loss functions used in OSTrack~\cite{ye2022Ostrack} (i.e., focal loss $\mathcal{L}_{focal}$, $L_1$ loss $\mathcal{L}_{L1}$, and GIoU loss $\mathcal{L}_{GIoU}$) and three knowledge distillation functions are used to optimize our visual tracker. Generally speaking, the overall loss can be denoted as: 
\begin{align}
    \label{lossFunction} 
    & \mathcal{L}_{total} = \lambda_1 \mathcal{L}_{focal} + \lambda_2 \mathcal{L}_{L1} + \lambda_3 \mathcal{L}_{GIoU} + \\ 
    &~~~~~~~~~~~~~~~~ \eta_1 \mathcal{L}_{simKD} + \eta_2 \mathcal{L}_{featKD} + \eta_3 \mathcal{L}_{resKD}   \nonumber 
\end{align}
For the first three loss functions for tracking, we refer the readers to check OSTrack~\cite{ye2022Ostrack} for better understanding. In the following paragraphs, we will describe the hierarchical knowledge distillation loss functions in detail.

\noindent 
\textbf{Similarity Matrix based Distillation.~} 
The similarity matrix computed in the multi-head self-attention layers incorporates abundant long-range and cross-modal relation information. In this work, we exploit the knowledge transfer from the similarity matrix learned by the teacher Transformer to the student Transformer. Specifically, we denote the similarity matrix of the $i^{th}$ teacher Transformer layer as $S_t^i \in \mathbb{R}^{640 \times 640}$. The similarity matrix of the $j^{th}$ student Transformer is denoted as $S_s^j \in \mathbb{R}^{320 \times 320}$. In addition to tracking loss functions, the learning of similarity matrix $S_s^j$ also depends on distilling loss $\mathcal{L}_{simKD}$ as follows, 
\begin{equation}
    \label{similarityMatrixDistill} 
    \mathcal{L}_{simKD} = \sum (S_s^j - S_t^i)^2. 
\end{equation}

\begin{table*}
\center
\footnotesize
\caption{Event camera based datasets for visual tracking. \# denotes the number of corresponding items.} 
\label{benchmarkList}
\resizebox{\textwidth}{!}{ 
\begin{tabular}{l|cccccccccccccccc}
\hline
\textbf{Datasets}    &\textbf{Year}	&\textbf{\#Videos}  &\textbf{\#Frames} &\textbf{\#Class}    &\textbf{\#Att} &\textbf{\#Resolution} &\textbf{Aim}   &\textbf{Absent}  &\textbf{Frame} &\textbf{Reality}  &\textbf{Public}  \\ 
\hline
\textbf{VOT-DVS}     &2016    &60           &-      &-    	 &-    &$240 \times 180$  &Eval  &\xmark   &\xmark     &\xmark     &\cmark     \\
\textbf{TD-DVS}        &2016     &77          &-      &-    	 &-  &$240 \times 180$  &Eval  &\xmark    &\xmark     &\xmark   &\cmark     \\
\textbf{Ulster}  &2016      &1     &9,000  		   &-    	 &-    &$240 \times 180$ 
 &Eval   &\xmark  &\xmark     &\cmark     &\xmark    		\\
\textbf{EED} &2018     &7     &234   &  -  	 &-   &$240 \times 180$   &Eval  &\xmark   &\xmark      &\cmark     &\cmark     \\
\hline 
\textbf{FE108}	&2021		&$108$     &208,672  	& 21  & 4     &$346 \times 260$     &Train \& Eval   &\xmark     &\xmark     &\cmark       &\cmark    \\
\textbf{VisEvent}  	&2021     &$820$       &371,127      	&  -  	&\textbf{17}     &$346 \times 260$  &Train \& Eval      &\cmark &\cmark          &\cmark           &\cmark     \\
\textbf{COESOT}  	&2022     &\textbf{1354}     &478,721       &\textbf{90}    	&\textbf{17}   &$346 \times 260$    & Train \& Eval  &\cmark &\cmark  &\cmark   &\cmark     \\
\hline 
\textbf{EventVOT}    &2023     &1141     &\textbf{569,359}       &19    	&14   &\textbf{1280} $\times$ \textbf{720}    & Train \& Eval  &\cmark &\xmark  &\cmark   &\cmark     \\ 
\hline
\end{tabular}
}
\end{table*}

\noindent 
\textbf{Feature based Distillation.~}
The feature distillation from the robust and powerful teacher Transformer network is the second strategy. We denote the token representations of the teacher and student network as $F_t$ and $F_s$. Then, the distilling loss between them can be represented as, 
\begin{equation}
    \label{featDistill} 
    \mathcal{L}_{featKD} = 
    \|F_t -F_s\|^2_F 
\end{equation}

\noindent 
\textbf{Response based Distillation.~} 
The response maps output from tracking networks are used for target object localization. Obviously, if we can directly mimic this response map $R_t$, the obtained tracking results will be better. In this paper, the weighted focal loss function~\cite{law2018cornernet} is adopted to achieve this target. We denote the ground truth target center and the corresponding low-resolution equivalent as $\hat{p}$ and $\bar{p} = [ \bar{p}_x, \bar{p}_y ]$, respectively. The Gaussian kernel is used to generate the ground truth heatmap $\hat{\bf{P}}_{xy} = exp (- \frac{(x-\bar{p}_x)^2 + (y-\bar{p}_y)^2}{2\delta^2_p})$, where $\delta$ denotes the object size-adaptive standard deviation~\cite{law2018cornernet}. Thus, the Gaussian Weighted Focal (GWF) loss function can be formulated as: 
\begin{equation}
\label{GaussianWFocalLoss} 
\small 
\mathcal{L}_{GWF} = - \sum_{xy} 
\begin{cases}
(1-\textbf{P}_{xy})^\alpha log(\textbf{P}_{xy}),   & if~\hat{\textbf{P}}_{xy} = 1   \\
(1-\hat{\textbf{P}}_{xy})^\beta(\textbf{P}_{xy})^\alpha log(1-\textbf{P}_{xy}),   &otherwise 
\end{cases}
\end{equation}
where $\alpha$ and $\beta$ are two hyper-parameters and are set to 2 and 4 respectively in our experiments, as suggested in OSTrack~\cite{ye2022Ostrack}. In our implementation, we normalize the response maps of both the teacher and student networks by dividing them by a temperature coefficient $\tau$ (empirically set to 2), followed by inputting them into the focal loss for response distillation, i.e., $\mathcal{L}_{resKD} = \mathcal{L}_{GWF}(R_s/\tau, R_t/\tau )$.

\begin{figure*}[!htp]
\center
\includegraphics[width=7in]{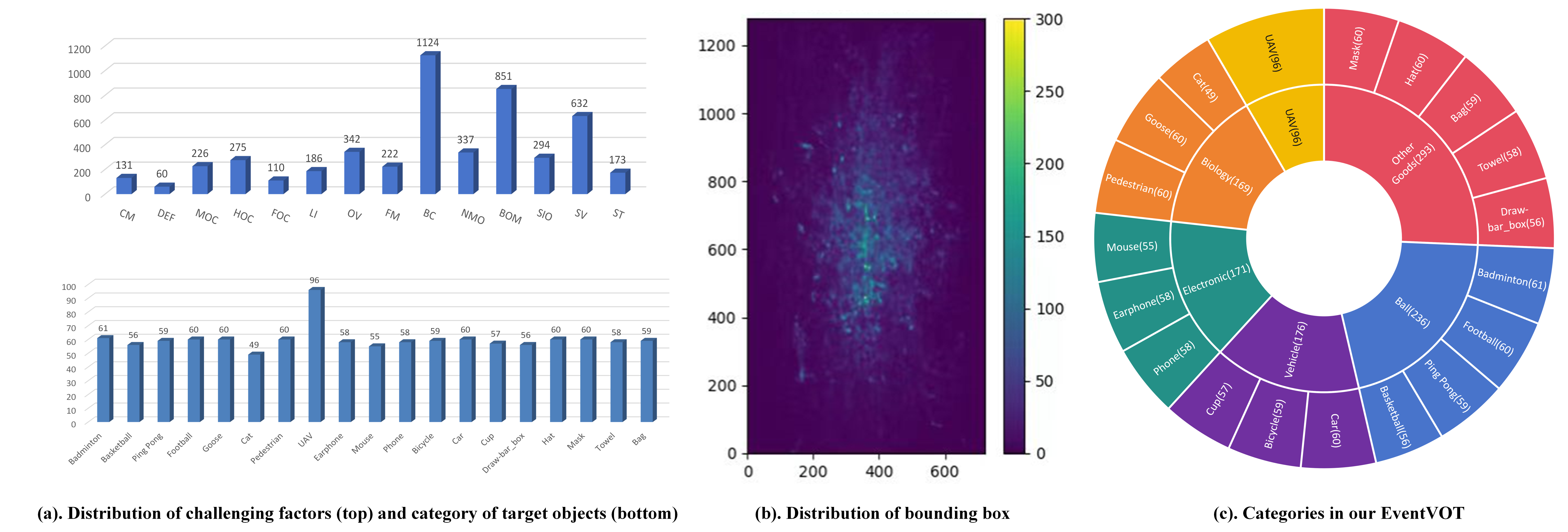}
\caption{Distribution of challenging factors, category of the target object, and bounding box.} 
\label{datasetInfo}
\end{figure*}

\section{EventVOT Dataset} 

In this paper, we propose a large-scale high-definition event-based visual tracking dataset. We will introduce the protocols for data collection and annotation, statistical analysis, and benchmarked visual trackers in the following subsections, respectively. Some representative samples are visualized in Fig.~\ref{eventvotExamples} and our \textcolor{magenta}{demo video} \footnote{\url{https://youtu.be/FcwH7tkSXK0}}.

\subsection{Criteria for Collection and Annotation} 
\begin{figure*}[!htp]
\center
\includegraphics[width=6.8in]{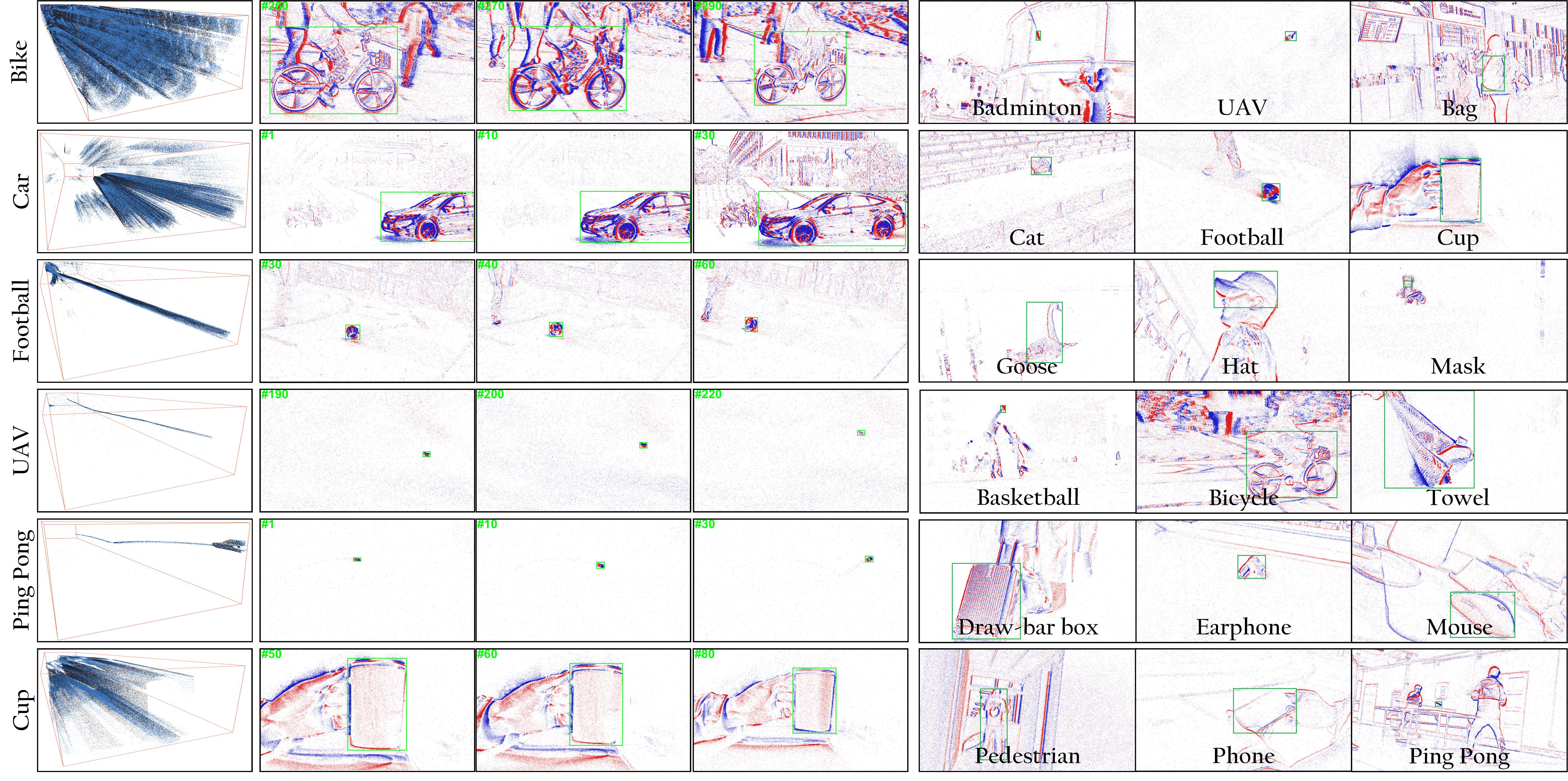}
\caption{Representative samples of our proposed EventVOT dataset. The $1^{th}$ column is the 3D event point stream and the $2^{th}$-$4^{th}$ columns are sampled event images. $5^{th}$-$7^{th}$ columns are more samples of our EventVOT dataset.}   
\label{eventvotExamples}
\end{figure*}

To construct a dataset with a diverse range of target categories, as shown in Fig.~\ref{eventvotExamples}, capable of reflecting the distinct features and advantages of event tracking, this paper primarily considered the following aspects during data collection. 
\emph{1). Diversity of target categories:} Many common and meaningful target objects are considered, including UAVs, pedestrians, vehicles, ball sports, etc. 
\emph{2). Diversity of data collection environments:} The videos in our dataset are recorded in day and night time, and involved venue information includes playgrounds, indoor sports arenas, main streets and roads, cafeteria, dormitory, etc. 
\emph{3). Recorded specifically for event camera characteristics:} Different motion speeds, such as high-speed, low-speed, momentary stillness, and varying light intensity, etc. 14 challenging factors are reflected by our EventVOT dataset. 
\emph{4). High-definition, wide-field event signals:} The videos are collected using a Prophesee EVK4–HD event camera, which outputs event stream with $1280 \times 720$. This high-definition event camera excels in supporting pure event-based object tracking, thereby avoiding the influences of the RGB cameras and showcasing its features and advantages in various aspects such as high-speed, low-light, low-latency, and low-power consumption.   
\emph{5). Data annotation quality:} All data is annotated by a professional data annotation company and has undergone multiple rounds of quality checks and iterations to ensure the accuracy of the annotations. For each event stream, we first stack into a fixed number (499 in our case) of event images for annotation. 
\emph{6). Data size:} Collect a sufficiently large dataset to train and evaluate robust event-based trackers. 
A comparison between the newly proposed dataset and existing tracking datasets is illustrated in Table~\ref{benchmarkList}.

\subsection{Statistical Analysis} 
In the EventVOT dataset, we define 14 challenging factors and involve 19 classes of target objects. The number of videos corresponding to these attributes and categories is visualized in Fig.~\ref{datasetInfo} (a, c). We can find that BC, BOM, and SV are top-3 major challenges which demonstrates that our dataset is relatively challenging. The balance between different categories is also well-maintained, with the number of samples roughly distributed between 50 to 60. Among them, UAVs (Unmanned Aerial Vehicles) are a special category of targets, with a total count of 96. The distribution of the center points of the annotated bounding boxes is visualized in Fig.~\ref{datasetInfo} (b).

\subsection{Benchmarked Trackers} 
To build a comprehensive benchmark dataset for event-based visual tracking, we consider the following visual trackers: 
\textbf{1). Siamese or Discriminate trackers:} DiMP50~\cite{goutam2019Dimp}, PrDiMP~\cite{martin2020PrDimp}, 
KYS~\cite{bhat2022SKys}, ATOM~\cite{martin2019Atom}, 
\textbf{2). Transformer trackers:} OSTrack~\cite{ye2022Ostrack}, TransT~\cite{chen2021transt}, SimTrack~\cite{chen2022SimTrack}, AiATrack~\cite{gao2022AIa}, STARK~\cite{yan2021Stark}, ToMP50~\cite{mayer2022Tomp}, MixFormer~\cite{cui2022Mixformer}, TrDiMP~\cite{wang2021TrDiMP}. 
Note that, we re-train these trackers using their default settings on our training dataset, instead of directly testing on the testing subset. Our EventVOT dataset is split into training/validation/testing subset which contains 841, 18, and 282 videos, respectively. We believe that these retrained tracking algorithms can play a crucial role in future comparisons of their performance.

\begin{table}
\center
\small  
\caption{Description of 14 attributes in our EventVOT dataset.} 
\label{AttributeList}
\resizebox{0.45\textwidth}{!}{ 
\begin{tabular}{l|lcccccccccccccc}
\hline \toprule [0.5 pt]
\textbf{Attributes}    &\textbf{Description}  \\ 
\hline
\textbf{01. CM}   	    	&Abrupt motion of the camera \\	
\textbf{02. MOC}   	    &Mildly occluded \\	
\textbf{03. HOC}   	    &Heavily occluded \\
\textbf{04. FOC}   	    &Fully occluded \\
\textbf{05. DEF}   	    &The target is deformable \\	
\textbf{06. LI}   	    	&Low illumination \\ 
\textbf{07. OV}   	    	&The target completely out of view \\ 
\textbf{08. SV}   	    	&Scale variation  \\
\textbf{09. BC}   	    	&Background clutter  \\
\textbf{10. FM}   	    	&Fast motion  \\
\textbf{11. NMO}   	    &No motion  \\
\textbf{12. BOM}         &Influence of background object motion \\ 
\textbf{13. SIO}         &Similar interferential object \\ 
\textbf{14. ST}          &Small target \\ 
\hline \toprule [0.5 pt]
\end{tabular}} 
\end{table}

\section{Experiment}

\subsection{Dataset and Evaluation Metric} 
In addition to our newly proposed \textbf{EventVOT} dataset, we also compare with other SOTA visual trackers on existing event-based tracking datasets, including \textbf{FE240hz}~\cite{zhang2021fe108}, \textbf{VisEvent}~\cite{wang2021visevent}, and \textbf{COESOT}~\cite{tang2022coesot} dataset. 
A brief introduction to these event-based tracking datasets is given below. 

$\bullet$ \textbf{FE240hz dataset}: It is collected using a gray-scale DVS346 event camera which contains 71 training videos and 25 testing videos. More than 1132K annotations on more than 143K images and corresponding events are provided. It considers different degraded conditions for tracking, such as motion blur and high dynamic range. 

$\bullet$ \textbf{VisEvent dataset}: It is the first large-scale frame-event tracking dataset recorded using a color DVS346 event camera. A total of 820 videos are collected in both indoor and outdoor scenarios. Specifically, the authors split these videos into a training subset and a testing subset which contain 500 and 320 videos, respectively. More details can be found on GitHub \url{https://github.com/wangxiao5791509/VisEvent_SOT_Benchmark}. 

$\bullet$ \textbf{COESOT dataset}: It is a category-wide RGB-event-based tracking dataset that contains 90 categories and 1354 video sequences (478,721 RGB frames). 17 challenging factors are formally defined in this dataset. The training and testing subset contains 827 and 527 videos, respectively. Please refer to the following GitHub for more details \url{https://github.com/Event-AHU/COESOT}.

For the evaluation metrics, we adopt the widely used \textbf{Precision Rate (PR)}, \textbf{Normalized Precision Rate (NPR)}, and \textbf{Success Rate (SR)}. The efficiency is also an important metric for a practical tracker, in this work, we adopt \textbf{FPS (Frames Per Second)} to measure the speed of a tracker.

\subsection{Implementation Details} 
The training of our tracker can be divided into two stages. We first pre-train the teacher Transformer with multimodal inputs for 50 epochs. The learning rate is 0.0001, weight decay is 0.0001, and batch size is 32. 
Then, the hierarchical knowledge distillation strategy is adopted for the training of the student Transformer network. The learning rate, weight decay, and batch size are set as 0.0001, 0.0001, and 32, respectively. 
The AdamW~\cite{loshchilov2018adamw} is selected as the optimizer. 
Our code is implemented using Python based on PyTorch~\cite{paszke2019pytorch} framework and the experiments are conducted on a server with CPU Intel(R) Xeon(R) Gold 5318Y CPU @2.10GHz and GPU RTX3090.

\subsection{Comparison on Public Benchmarks} 

\noindent 
\textbf{Results on FE240hz Dataset.~} 
As shown in Table~\ref{FE240table}, our baseline OSTrack achieves 57.1/89.3 on the SR/PR metric, meanwhile, ours are 59.8/92.2 which is significantly better than the baseline method. Our tracker also beats other SOTA trackers including event-based trackers (e.g., STNet and EFE), and Transformer trackers (like TransT, STARK) by a large margin. These results fully validated the effectiveness of our proposed hierarchical knowledge distillation strategy for event-based tracking.

\begin{table}
\center
\small     
\caption{Experimental results (SR/PR) on FE240hz dataset.} 
\label{FE240table}
\resizebox{\columnwidth}{!}{
\begin{tabular}{cccccccccc}
\hline 
\textbf{STNet}  &\textbf{TransT}  &\textbf{STARK}    &\textbf{PrDiMP}  &\textbf{EFE}    &\textbf{SiamFC++}    \\ 
58.5/89.6      &56.7/89.0        &55.4/83.7       &55.2/86.8        &55.0/83.5       &54.5/85.3 \\ 
\hline 
\textbf{DiMP}   &\textbf{ATOM}    &\textbf{Ocean}    &\textbf{SiamPRN}    &\textbf{OSTrack}    &\textbf{Ours}     \\ 
53.4/88.2      &52.8/80.0         & 50.2/76.4        &41.6/75.5           &57.1/89.3          &59.8/92.2    \\         
\hline 
\end{tabular}
}
\end{table}

\begin{table}
\center
\small     
\caption{Results on VisEvent dataset. EF and MF are short for early fusion and middle-level feature fusion.} 
\label{Viseventtable}
\begin{tabular}{c|l|ccc}
\hline 
&\textbf{Trackers}   &\textbf{SR}  &\textbf{PR} &\textbf{NPR} \\   
\hline 
\multirow{9}{*}{\rotatebox{90}{\textbf{RGB + Event Input}}}
&\textbf{CEUTrack} &{64.89}   &{69.06} &{73.81} \\
&\textbf{LTMU (EF)} &  60.10  & 66.76 & 69.78\\
&\textbf{PrDiMP (EF)}&  57.20   & 64.47 & 67.02\\
&\textbf{CMT-MDNet (MF)} & 57.44   & 67.20  & 69.78\\
&\textbf{ATOM (EF) }  & 53.26 & 60.45 & 63.41\\
&\textbf{SiamRPN++ (EF) } & 54.11   & 60.58 & 64.72\\
&\textbf{SiamCAR (EF) } & 52.66 & 58.86  & 62.99\\
&\textbf{Ocean (EF) }  & 43.56  & 52.02 & 54.21\\
&\textbf{SuperDiMP (EF) } & 36.21   & 46.99 &42.84 \\
\hline   
\multirow{5}{*}{\rotatebox{90}{\textbf{Event Input}}}
&\textbf{STNet (Event-Only)}  & 39.7   & 49.2 &- \\
&\textbf{TransT (Event-Only)}  & 39.5   & 47.1  &- \\
&\textbf{STARK (Event-Only)}  & 34.8  & 41.8  &- \\
&\textbf{OSTrack (Event-Only)}  & 34.5    & 48.9   & 38.5 \\
&\textbf{Ours (Event-Only)}  &37.3 &52.5    &41.0   \\
\hline 
\end{tabular}
\end{table}

\noindent 
\textbf{Results on VisEvent Dataset.~} As shown in Table~\ref{Viseventtable}, we report the tracking results on the VisEvent dataset and compare them with multiple recent strong trackers. Specifically, our baseline OSTrack~\cite{ye2022Ostrack} achieves $34.5, 48.9, 38.5$ on SR, PR, and NPR, respectively, meanwhile, ours are $37.3, 52.5, 41.0$ on these metrics. These results demonstrate that our proposed hierarchical knowledge distillation strategy can enhance the event-based tracking results by learning from multimodal input data. Compared with other Transformer based trackers, such as the STARK~\cite{yan2021Stark}, we can find that our results are much stronger than this tracker, with an improvement of +2.5 and +10.7 on SR and PR. We also beat the STNet and TransT on the PR metric, which fully validated the effectiveness of our proposed strategy for event-based tracking.

\noindent 
\textbf{Results on EventVOT Dataset.}  
As shown in Table~\ref{EventVOT_auc}, we re-train and report multiple SOTA trackers on the EventVOT dataset. We can find that our baseline tracker OSTrack achieves $55.4, 56.4, 65.2$ on the SR, PR, and NPR, respectively. When adopting our proposed hierarchical knowledge distillation framework in the training phase, these results can be improved to $57.0, 57.3, 66.5$ which fully validated the effectiveness of our proposed method for event-based tracking. Our results are also better than other SOTA trackers, including the Siamese trackers and Transformer based trackers (STARK, MixFormer, PrDiMP, etc.).

\begin{figure*}
\center
\includegraphics[width=7in]{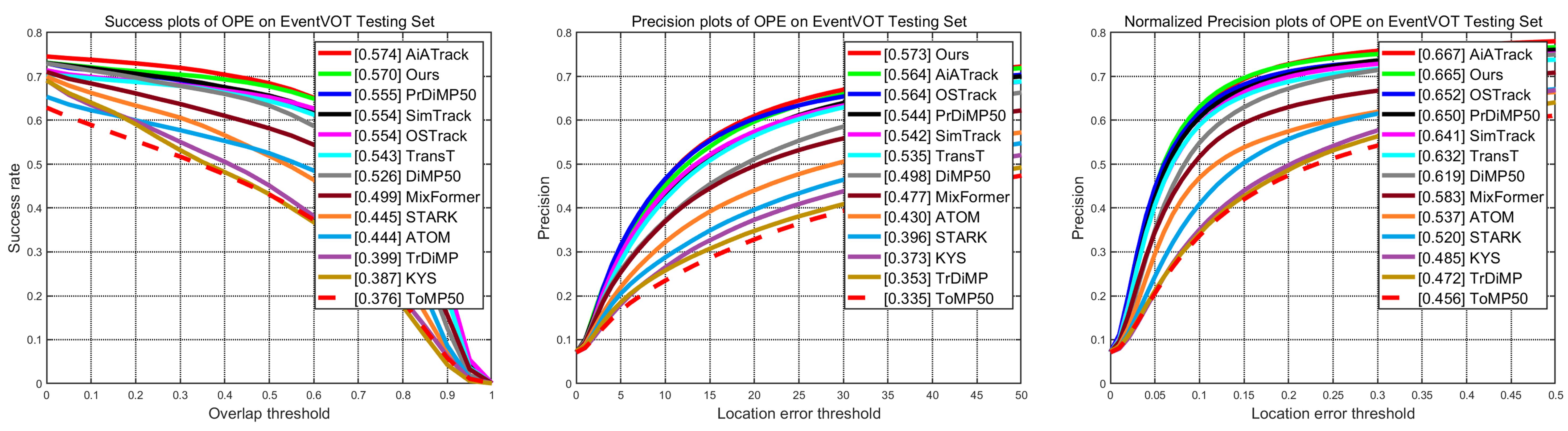}
\caption{Visualization of tracking results of our proposed EventVOT dataset.}  
\label{PRSRNPRfig}
\end{figure*}

\begin{table}
\center
\small   
\caption{Overall tracking performance on EventVOT dataset. } 
\label{EventVOT_auc}
\resizebox{\columnwidth}{!}{ 
\begin{tabular}{l|c|ccc|cc}
\hline 
\textbf{Trackers} & \textbf{Source}   & \textbf{SR}  &\textbf{PR}   &\textbf{NPR}  &\textbf{Params}  &\textbf{FPS}\\
\hline
\textbf{ Ours}      &--  &\ 57.0   &\ 57.3   &\ 66.5   &\ 92.1   &\ 105  \\ 
\textbf{ TrDiMP } & CVPR21     &\ 39.9   &\ 35.3   &\ 47.2   &\ 26.3   &\ 26   \\ 
\textbf{ ToMP50  }   &  CVPR22   &\ 37.6   &\ 33.5   &\ 45.6   &\ 26.1   &\ 25  \\ 
\textbf{ OSTrack }   &  ECCV22   &\ 55.4  &\ 56.4   &\ 65.2   &\ 92.1   &\ 105  \\
\textbf{ AiATrack   }   &  ECCV22     &\ 57.4   &\ 56.4   &\ 66.7   &\ 15.8   &\ 38  \\ 
\textbf{ STARK   }   &  ICCV21     &\ 44.5   &\ 39.6  &\ 52.0   &\ 28.1   &\ 42  \\ 
\textbf{ TransT   }   &  CVPR21     &\ 54.3  &\ 53.5  &\ 63.2   &\ 18.5   &\ 50  \\ 
\textbf{ DiMP50   }  &  ICCV19       &\ 52.6   &\ 49.8   &\ 61.9   &\ 26.1  &\ 43  \\
\textbf{ PrDiMP   }  &  CVPR20       &\ 55.5   &\ 54.4   &\ 65.0   &\ 26.1   &\ 30  \\
\textbf{ KYS   }   &   ECCV20         &\ 38.7   &\ 37.3   &\ 48.5   &\ --   &\ 20  \\ 
\textbf{ MixFormer   }   & CVPR22     &\ 49.9   &\ 47.7   &\ 58.3   &\ 35.6   &\ 25  \\
\textbf{ ATOM   }   & CVPR19     &\ 44.4   &\ 43.0   &\ 53.7   &\ 8.4   &\ 30  \\
\textbf{ SimTrack   }   & ECCV22     &\ 55.4   &\ 54.2  &\ 64.1   &\ 57.8   &\ 40  \\ 
\hline
\end{tabular}
}
\end{table}

\noindent 
\textbf{Results on COESOT Dataset.}  
As shown in Table~\ref{COESOT_results}, we report our tracking results on the large-scale RGB-Event tracking dataset COESOT. Note that, the compared baseline methods are re-trained on the training subset of COESOT using their default settings and hyper-parameters to achieve a relatively fair comparison. It is easy to find that our baseline OSTrack achieves $50.9, 57.8, 56.7$ on the SR, PR, and NPR metrics, meanwhile, we obtain $52.3, 59.0, 58.0$ which are significantly better than theirs. Our tracking results are also better than most of the compared trackers, including TransT, AiATrack, MixFormer, etc. These experimental results fully demonstrate the effectiveness of our proposed hierarchical knowledge distillation from multi-modal to event-based tracking networks.

\begin{table}
\center
\small   
\caption{Overall tracking performance on COESOT dataset. } 
\label{COESOT_results}
\begin{tabular}{l|c|ccccc}
\hline 
\textbf{Trackers} & \textbf{Source}   & \textbf{SR}  &\textbf{PR}   &\textbf{NPR} \\
\hline
\textbf{ Ours}    &-     &52.3       &59.0       &58.0           \\ 
\textbf{ TrDiMP }   & CVPR21     &50.7       &56.9       &55.2           \\ 
\textbf{ ToMP50  }   &  CVPR22   &46.3       &52.9       &52.5           \\ 
\textbf{ OSTrack }   &  ECCV22   &50.9       &57.8       &56.7           \\ 
\textbf{ AiATrack   }   &ECCV22   &51.3       &57.9       &56.2           \\ 
\textbf{ STARK   }   &  ICCV21    &40.8       &44.9       &44.4           \\ 
\textbf{ TransT   }   &  CVPR21     &45.6       &51.4       &50.4           \\ 
\textbf{ DiMP50   }  &  ICCV19     &53.8       &61.7       &60.3           \\ 
\textbf{ PrDiMP   }  &  CVPR20     &47.5       &55.1       &54.0           \\ 
\textbf{ KYS   }   &   ECCV20      &42.6       &50.6       &49.7           \\ 
\textbf{ MixFormer   }   & CVPR22   &44.4       &49.4      &48.5           \\ 
\textbf{ ATOM   }   & CVPR19    &42.1       &48.0        &48.1           \\ 
\textbf{ SimTrack   }   & ECCV22  &48.3       &53.5       &52.9           \\  
\hline
\end{tabular}
\end{table}

\subsection{Ablation Study}

\begin{table}
\center
\small     
\caption{Component Analysis on COESOT and EventVOT.} 
\label{CAResults} 
\resizebox{\columnwidth}{!}{
\begin{tabular}{c|cccc|cc} 		
\hline 
\textbf{No.}  & \textbf{Base} &\textbf{SKD} &\textbf{FKD}  &\textbf{RKD}        &\textbf{COESOT}   &\textbf{EventVOT} \\
\hline 
1 &\cmark   &          &         &         &\ 57.8/50.9         &\ 56.4/55.4        \\
2 &\cmark   &\cmark    &         &         &\ 58.4/51.6         &\ 56.8/56.5         \\
3 &\cmark   &    &\cmark         &         &\ 59.0/52.1         &\ 56.6/56.4         \\
4 &\cmark   &   &    &\cmark         &\ 58.3/51.5         &\ 56.6/56.2          \\
5 &\cmark   &\cmark    &\cmark   &\cmark   & \ 59.0/52.3      &\ 57.3/57.0          \\  
\hline
\end{tabular} } 
\end{table}

\noindent 
\textbf{Analysis on Hierarchical Knowledge Distillation.~}   
In this section, we will isolate each distillation strategy for individual experimentation to assess its impact on the final tracking performance. On the COESOT dataset, we take the RGB and event image as the input of teacher network and feed the event data only into the student tracker. For the EventVOT dataset, we stack the event stream into images and voxels and conduct hierarchical knowledge distillation based on multi-view settings. As shown in Table~\ref{CAResults}, the base denotes the tracker is trained using three tracking loss functions only as OSTrack and CEUTrack do. We can find that it achieves $57.8/50.9$, and $56.4/55.4$ on the COESOT and EventVOT datasets, respectively. When introducing new distillation loss like similarity-based, feature-based, and response-based distillation functions, the results are all improved in both settings. Note that, the feature-based distillation works better on the COESOT in contrast to the EventVOT dataset. When all these distillation strategies are used, better tracking performance can be obtained on multi-modal and multi-view settings. On the basis of all these experiments, we can draw the conclusion that all the proposed hierarchical knowledge distillation strategy contributes to event-based tracking.

\begin{figure}
\center
\includegraphics[width=3.3in]{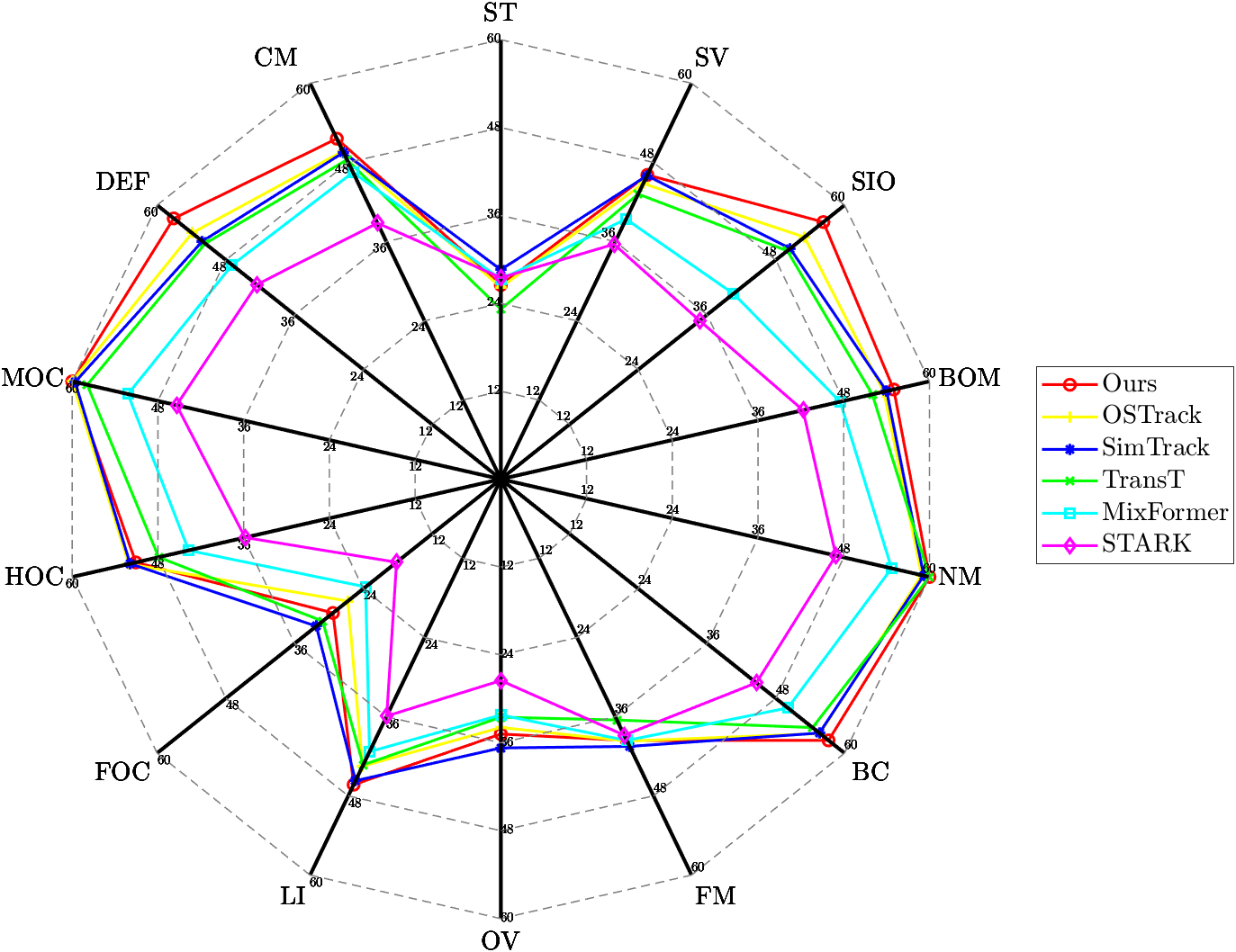}
\caption{Tracking results under each challenging factor.} 
\label{attributeResults}
\end{figure}

\noindent 
\textbf{Analysis on Tracking in Specific Challenging Environment.~}  
In this work, our proposed EventVOT dataset reflects 14 core challenging factors in the tracking task. As shown in Fig.~\ref{attributeResults}, we report the results of our tracker and other state-of-the-art trackers under each challenging scenario. We can find that our proposed tracker achieves better performance when facing attributes like DEF (Deformation), CM (Camera motion), SIO (Similar interferential object), BC (Background clutter), etc. We also achieve similar tracking results in other attributes which demonstrate that our proposed hierarchical knowledge distillation strategy works well for transferring knowledge from multi-modal/multi-view data to event-based tracker.

\noindent 
\textbf{Analysis on Different Event Representation.~}   
In this part, we conduct tracking with multiple representations of event data and analyze the influences of different event representations. Specifically, the event image, event voxel, and time surface are considered, as shown in Table~\ref{event_repr}. We can find that the event voxel based tracking performs worse than others on our high-resolution event stream. We think this might be because the feature representation of voxels requires careful design.

\begin{table}
\center
\small     
\caption{Ablation studies on event representation, loss functions, align methods, and the number of Transformer layers on EventVOT and COESOT dataset.} 
\label{event_repr}
\resizebox{\columnwidth}{!}{ 
\begin{tabular}{l|lll}
\hline 
\textbf{\#(EventVOT). Input Data}    &\textbf{SR}   & \textbf{PR}  & \textbf{NPR}  \\
\hline
\text{1. Event Frames }     &57.0   &57.3   &66.5  \\
\text{2. Event Voxels }     &8.6    &7.8     &12.4  \\
\text{3. Event Time Surface} &53.3   &52.3 &62.7  \\
\text{4. Event Reconstruction Images} &54.5  &55.8 &63.5  \\
\hline 
\textbf{\#(COESOT). Loss for Feature-level KD}    &\textbf{SR}   & \textbf{PR}  & \textbf{NPR}  \\
\text{5. MSE loss }     &52.1    &59.0    &57.9       \\ 
\text{6. L2 loss }      &51.9    &58.7    &57.5       \\
\text{7. L1 loss }      &51.6    &58.0    &56.9       \\
\text{8. KLD loss }     &50.2    &56.5    &55.3       \\
\hline 
\textbf{\#(COESOT). Align Method for Distillation}    &\textbf{SR}   & \textbf{PR}  & \textbf{NPR}  \\
\text{9. Repeat }                   &52.1      &59.0      &57.9       \\ 
\text{10. Reshape \& Resize }        &51.4      &57.7      &56.6       \\ 
\text{11. FC }                      &50.9      &57.3      &56.3       \\ 
\hline 
\textbf{\#(COESOT). Number of Former Layers}    &\textbf{SR}   & \textbf{PR}  & \textbf{NPR}  \\
\text{12. 12 layers }             &52.3      &59.0      &58.0       \\ 
\text{13. 8 layers }              &49.2      &55.5      &54.8       \\ 
\text{14. 4 layers }              &42.1      &46.0      &46.3       \\ 
\hline
\end{tabular}
}
\end{table}

\begin{figure*} 
\center
\includegraphics[width=7in]{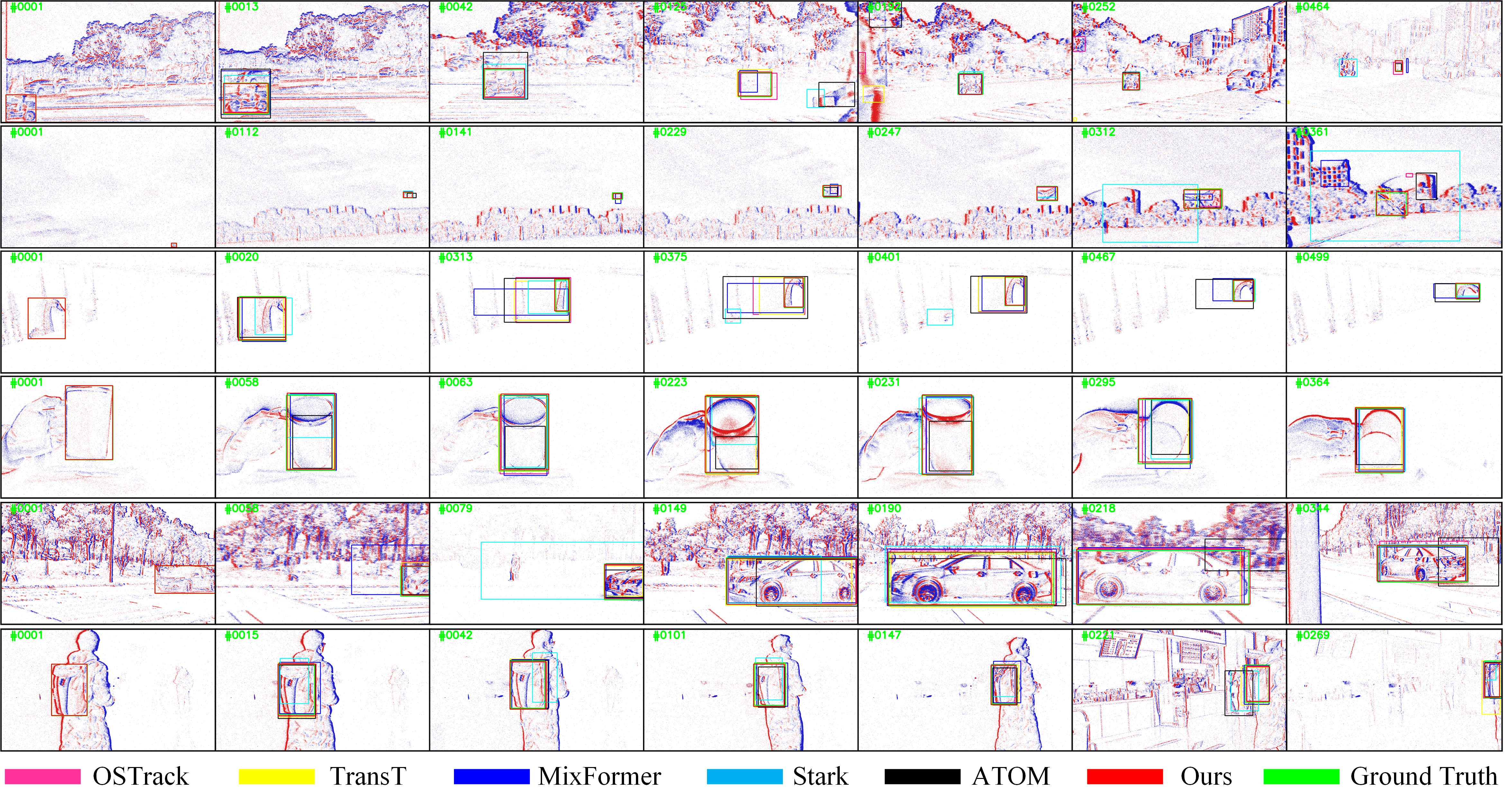}
\caption{Visualization of the tracking results of ours and other SOTA trackers.}  
\label{trackingResults}
\end{figure*}

\noindent 
\textbf{Analysis on Loss Function for Distillation.~} When conducting knowledge distillation in our training phase, multiple loss functions can be selected, such as L1, L2, MSE (Mean Squared Error), and KLD (Kullback-Leibler Divergence) loss functions. In this part, we test the effectiveness of these loss functions based on feature-level knowledge distillation and report the tracking results on COESOT dataset. As shown in Table~\ref{event_repr}, we can find that MSE performs the best which achieves 52.1 and 59.0 on SR and PR metric.

\noindent 
\textbf{Analysis on Number of Transformer Layers.~} 
When conducting tracking using student Transformer networks, the accuracy and tracking speed are influenced by the number of Transformer layers. In this part, we set different layers to check their influences, i.e., 12, 8, and 4 layers. It is easy to find that more Transformer layers (more learnable parameters) will bring us better tracking results.

\noindent 
\textbf{Analysis on Align Methods for Distillation.~} In the training phase, the number of teacher Transformer networks is twice large as the student network. We also tried different alignment approaches to bridge this gap for knowledge distillation, including repeating, reshaping and resizing, and adjusting using a fully connected layer. We can find that simple repeat features of the student network perform the best for event-based tracking.

\begin{figure}
\center
\includegraphics[width=3.3in]{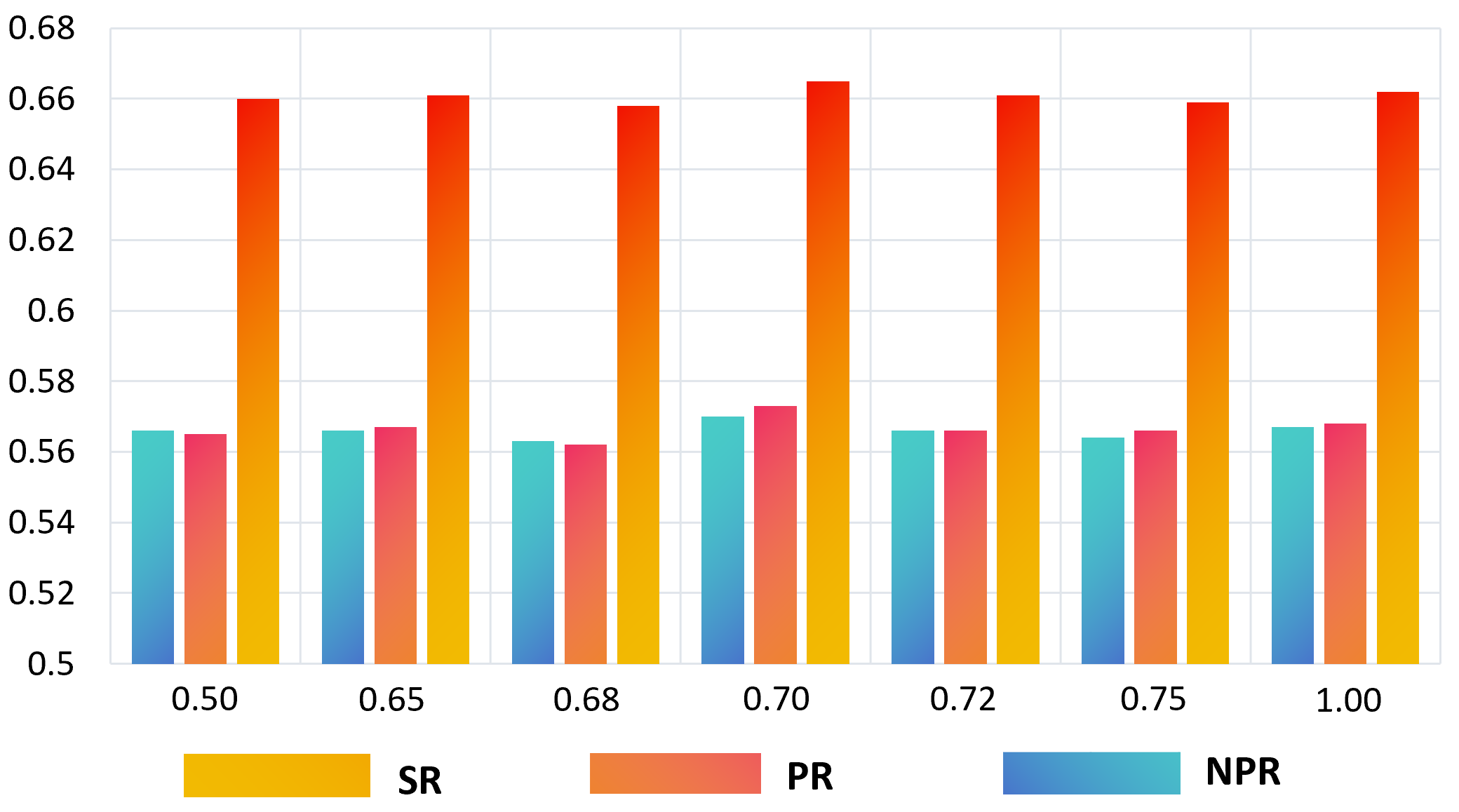}
\caption{Results with different tradeoff parameters for distillation.}  
\label{tradeoffKLD}
\end{figure}

\noindent 
\textbf{Analysis on Tradeoff Parameters for Distillation Strategies.~} In the hierarchical knowledge distillation phase, we set different tradeoff parameters to achieve better tracking performance. As feature-level distillation is widely exploited and also performs well on our dataset, therefore, we default set its weight as 1. For the similarity-level and response-level distillation, we experimentally set their weights as equal ones, including 0.50, 0.65, 0.68, 0.70, 0.72, 0.75, and 1. As shown in Fig.~\ref{tradeoffKLD}, better tracking results can be obtained if we set the weights as [1, 0.7, 0.7] which achieves 0.570, 0.573, 0.665 on SR, PR, and NPR, respectively.


\subsection{Visualization} 

\begin{figure}
\center
\includegraphics[width=3.3in]{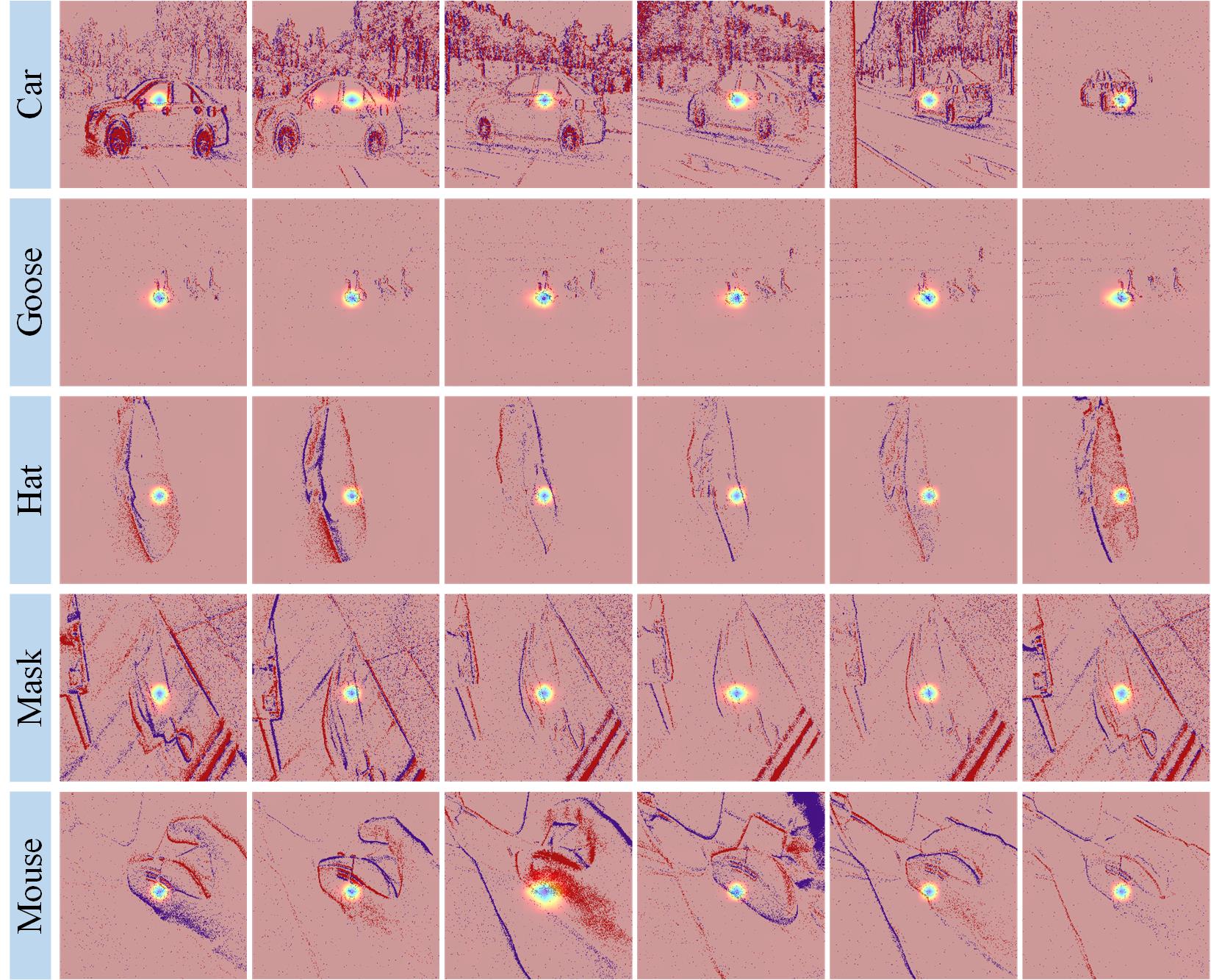}
\caption{Visualization of response maps predicted by our HDETrack.}  
\label{similarityAttentionMaps}
\end{figure}

In addition to the quantitative analysis mentioned above, we also conducted a visual analysis of the proposed tracking algorithm to provide readers with a better understanding of our tracking framework. 
As shown in Fig.~\ref{trackingResults}, we visualize the tracking results of ours and other SOTA trackers on the EventVOT dataset, including OSTrack, TransT, MixFormer, STARK, and ATOM. We can find that our tracking using event camera is an interesting and challenging task. These trackers perform well in simple scenarios, however, there is still significant room for improvement. 
Besides, we also provide the response maps of Transformer networks on multiple videos including \emph{Car, Goose, Hat, Mask,} and \emph{Mouse}. As shown in Fig.~\ref{similarityAttentionMaps}, we can find that the target object regions are highlighted which means our tracker focuses on the real targets accurately.

\section{Conclusion}  

In this paper, we propose a novel hierarchical knowledge distillation framework for event-based tracking. It formulates the learning of event trackers based on the teacher-student knowledge distillation framework. The teacher network takes the multi-modal or multi-view data as the input, meanwhile, the student network takes the event data for tracking. In the distillation phase, it simultaneously considers similarity-based, feature-based, and response-based knowledge distillation. 
To bridge the data gap, we also propose the first large-scale, high-resolution event-based tracking dataset, termed EventVOT. It contains 1141 video sequences and covers 19 categories of target objects like pedestrians, vehicles, UAVs, etc. More than 10 recent strong trackers are re-trained and evaluated on our dataset which makes the EventVOT becomes a more suitable benchmark for future works to compare. Extensive experiments on multiple datasets fully validated the effectiveness of our proposed hierarchical knowledge distillation strategy. 
In our future works, we will consider collecting more high-resolution event videos and pre-train a strong event-based tracker in a self-supervised learning manner.

{\small
\bibliographystyle{ieee_fullname}
\bibliography{reference}
}

\end{document}